\documentclass[]{lab}

\usepackage[utf8]{inputenc}
\usepackage[T1]{fontenc}
\usepackage{geometry}
\usepackage{amsmath,amssymb}  %
\usepackage{charter}

\usepackage[toc,page,header]{appendix}

\usepackage{minitoc}
\usepackage{cleveref} 
\usepackage{subcaption}
\usepackage{booktabs}
\usepackage{graphicx}
\usepackage{pgfplots}
\usepackage{pgfplotstable}
\usepackage{xcolor}
\usepackage{CJKutf8}
\usetikzlibrary{patterns}
\usepackage{float}
\usepackage{caption}
\usepackage{bm}
\usepackage{makecell} % in preamble
\usepackage{colortbl}

\usepackage{soul} %
\usepackage{algorithm}      %
\usepackage{algpseudocode}  %
\usepackage{parskip}

\definecolor{cvprblue}{rgb}{0.21,0.49,0.74}
\definecolor{fallbackgreen}{rgb}{130, 180, 102}
\definecolor{stopred}{rgb}{251, 225, 224}

\ifdefined\final
\usepackage[disable]{todonotes}
\else
\usepackage[textsize=tiny]{todonotes}
\fi

\newcommand{\ours}{\texttt{Vidu S1}\xspace}

\usepackage{natbib}
\usepackage{latexsym}

\usepackage{url}
\usepackage{amssymb}
\usepackage[utf8]{inputenc}
\usepackage{microtype}
\usepackage{booktabs}
\usepackage{pifont} 
\usepackage{multirow}
\usepackage{makecell}
\usepackage{paralist}
\usepackage{xspace}
\usepackage{color}
\usepackage{xcolor}
\usepackage{colortbl}
\usepackage{adjustbox}
\usepackage{hyperref} 
\usepackage[edges]{forest}
\usepackage{tikz} 
\usepackage{caption}
\usepackage{amsfonts}

\hypersetup{
    colorlinks,
    linkcolor={blue!80!black},
    citecolor={blue!80!black},
}
\tikzset{
    root/.style =             {align=center, text width=1cm, rounded corners=3pt, line width=0.3mm, fill=gray!10, draw=gray!80, font=\small},
    demographic/.style =         {align=center, text width=1.8cm, rounded corners=3pt, line width=0.3mm, fill=blue!10, draw=blue!80, font=\footnotesize},
    demographic_work/.style =    {align=center, text width=10cm, rounded corners=3pt, line width=0.3mm, fill=blue!10, draw=blue!0, font=\footnotesize},
    character/.style =         {align=center, text width=1.8cm, rounded corners=3pt, line width=0.3mm, fill=red!10, draw=red!80, font=\footnotesize},
    character_work/.style =    {align=center, text width=10cm, rounded corners=3pt, line width=0.3mm, fill=red!10, draw=red!0, font=\footnotesize},
    personalization/.style =           {align=center, text width=1.8cm, rounded corners=3pt, line width=0.3mm, fill=cyan!10, draw=cyan!80, font=\footnotesize},
    personalization_work/.style =      {align=center, text width=10cm, rounded corners=3pt, line width=0.3mm, fill=cyan!10, draw=cyan!0, font=\footnotesize},
    risk/.style =         {align=center, text width=1.8cm, rounded corners=3pt, line width=0.3mm, fill=orange!10, draw=orange!80, font=\footnotesize},
    risk_work/.style =    {align=center, text width=10cm, rounded corners=3pt, line width=0.3mm, fill=orange!10, draw=orange!0, font=\footnotesize},
}

\usepackage{CJK}

\renewcommand{\thefootnote}{}

\newtcolorbox{promptbox}[1][]{
  enhanced,
  breakable,
  colback=promptboxlightgray,
  colframe=promptboxblue!30,
  arc=8pt,
  boxrule=0.5pt,
  left=12pt,
  right=12pt,
  top=8pt,
  bottom=8pt,
  fonttitle=\bfseries,
  fontupper=\linespread{1.2}\selectfont,
  title=#1
}

\title{Vidu S1: A Real-Time Interactive Video Generation Model}

\author{{\fontsize{11}{11}\selectfont Jintao Zhang$^{*\dagger}$, Kai Jiang$^{*}$, Jintao Chen$^{*}$, Xu Wang$^{*}$, Yang Luo$^{*}$, Yuji Wang$^{*}$, Dechuang Chen$^{*}$, Jungang Li$^{*}$, Chengyang Ye, Marco Chen, Hongzhou Zhu, Min Zhao, Yuxuan Jiang, Zhengkun Huang, Chendong Xiang, Kaiwen Zheng, Haoxu Wang, Xiaohang Wang, Qi Jia, Xin Chen, Yimin Chen, Youhe Jiang, Fangcheng Fu, Zhijie Deng$^{\ddagger}$, Fan Bao$^{\ddagger}$, Jianfei Chen$^{\ddagger}$, Jun Zhu$^{\ddagger}$}}

\affiliation{Tsinghua University, Shengshu Technology\\\url{https://vidu.com/vidu-stream}}

\vspace{-10pt}
\abstract{
We introduce \ours, a real-time interactive video generation model supporting voice control of digital characters. Users can control video generation content at any moment through voice instructions. \ours supports infinite-length real-time video generation without blurring, drift, or visual distortion. Built with TurboDiffusion and TurboServe, \ours outputs 540p real-time videos at up to 42 FPS on regular consumer GPUs. Users can upload custom images of real people, anime, and pets, and choose different voice tones for personalized experiences. Experiments show that \ours achieves the best performance across all test metrics while fully meeting real-time inference requirements. A playable online demo is available at \url{https://vidu.com/vidu-stream}.
}
\begin{document}

\maketitle

\renewcommand{\thefootnote}{}
\footnotetext{*Core contributors, co-first authorship. $^\dagger$Project lead. $^\ddagger$Advisors.}

\renewcommand{\thefootnote}{\arabic{footnote}}

\begin{figure}[H]
    \vspace{-20pt}
    \centering
    \includegraphics[width=\linewidth]{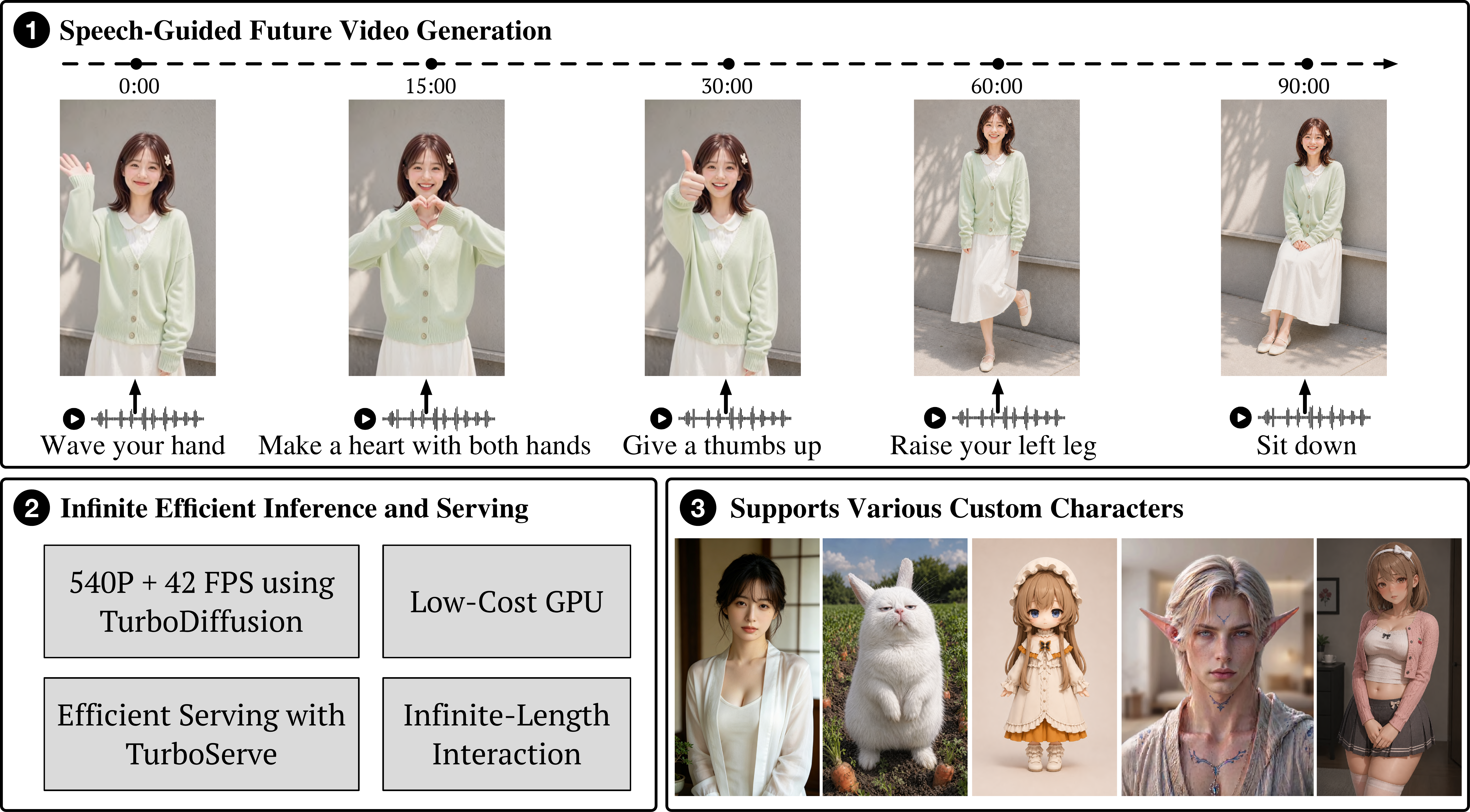}
    \caption{Overview of \ours. \ours supports speech-guided future control, efficient infinite-duration inference, and diverse customized characters for real-time interactive video generation.}
    \label{fig:teaser}
\end{figure}

\section{Introduction}

\paragraph{Background and Demand.}
Recent video generation models, such as Sora, Veo, Wan, and Seedance~\citep{videoworldsimulators2024,veo,wan2025wan,seedance2026seedance}, have shown strong ability in generating high-quality videos. However, most of them still follow an offline, one-shot generation paradigm: a user enters a prompt, waits for minutes or even tens of minutes, and receives the complete video only after generation finishes. 
During this process, the user can only passively wait to receive information and cannot actively initiate any interaction.
This limitation comes from the offline diffusion paradigm, where the model denoises the entire video synchronously over many steps and only produces an entire clean video at the end. Such a paradigm works well for offline content creation, but human visual entertainment is not limited to pre-generated videos. People also enjoy face-to-face communication, live streaming, games, talking with someone, and other interactive visual experiences, where content must respond immediately to the user. From a demand perspective, suppose that each user has an average demand of $\alpha \in [0,1]$ for real-time interactive visual content, e.g., $\alpha = 0.5$. Then the total demand scales with $\alpha \times N$, where $N$ is the number of users. In contrast, suppose that each user has an average demand of $\beta \in [0,1]$ for offline-generated visual content, e.g., $\beta = 0.5$. Since offline-generated videos can be replayed and shared, their generation demand scales more like $\beta \times N / m$, where $m$ is the average number of views per generated video. If we assume $\alpha \approx \beta$ and $m > 100$, then the demand for real-time interactive video generation is much greater than that for offline-generated videos.

\begin{center}
\textbf{Real-time interactive video generation is one of the most important future directions for industry.}
\end{center}
\vspace{-6pt}

\paragraph{Limitation.}
Despite the importance of this direction, existing autoregressive video generation models still have several key limitations. (1) Many methods~\citep{DF,kim2024fifo,henschel2025streamingt2v,song2025history} only change the generation process from offline generation to autoregressive generation, but they still do not support real-time user interaction during generation. (2) Most methods~\citep{liu2025rolling,sun2025ar,xie2025progressive,feng2025streamdiffusionv2} do not use speech as a direct and explicit control signal for future video content, making it difficult for users to control the generation process through natural spoken instructions. (3) Long-horizon video generation remains challenging because small errors can accumulate over time, causing drift, instability, and eventually visual collapse~\citep{yin2025slow,huang2026self}. As a result, many models~\citep{huang2026self,liu2025rolling,yin2025slow,zhu2026causal,zhao2026causal,yang2025longlive,teng2025magi,chen2025skyreels,zhang2025packing,gao2025wan} can only generate videos within a limited duration and cannot maintain stable generation over an open-ended stream. (4) Real-time video generation is not only a modeling problem. It also requires strong inference kernels, serving systems, scheduling strategies, and large-scale deployment infrastructure. Without such infrastructure, even a capable model may be too slow or too expensive for practical use.

\vspace{-1pt}
\paragraph{Vidu S1.}
To address these limitations, we introduce \ours, a real-time interactive video generation model designed for continuous user interaction. 
(1)\ours allows users to actively interact with the generation process at any moment, intervening in real time instead of merely specifying all controls before generation begins.
(2) \ours treats user speech as a direct and explicit control signal, so spoken instructions can continuously guide what the video should generate next.
(3) \ours supports stable long-horizon real-time generation by mitigating error accumulation during streaming generation, allowing the video to continue indefinitely without rapid drift or collapse.
(4) It is also built together with an efficient inference and serving stack, including TurboDiffusion~\citep{zhang2025turbodiffusion} and TurboServe~\citep{jiang2026turboserve}, which enables low-cost real-time inference on commodity GPUs at 540p resolution and 42 FPS. 

\vspace{-3pt}
\paragraph{Contribution.}
We summarize our contributions as follows.

\begin{enumerate}
\item We introduce \ours, a real-time interactive video generation model that supports continuous user interaction and enables explicit speech control over future video content during generation.

\item \ours supports indefinitely long video generation with high quality and consistency. By mitigating error accumulation during streaming, it allows video streams to run without drift or collapse.

\item \ours achieves real-time video generation at 540p resolution and 42 FPS. With an efficient inference and serving stack, \ours makes high-quality real-time generation practical on low-cost GPUs.

\item Experiments on Vidu-StreamBench and HDTF demonstrate that \ours combines leading quality and real-time interactivity, achieving the best CSIM (0.9192), Sync-D (7.847), and DOVER (0.5660), strong human preference while supporting speech-guided control and stable long-horizon streaming.

\end{enumerate}

\section{Method}

\subsection{Data Preparation}
\begin{figure}[H]
    \centering
    \includegraphics[width=1.0\linewidth]{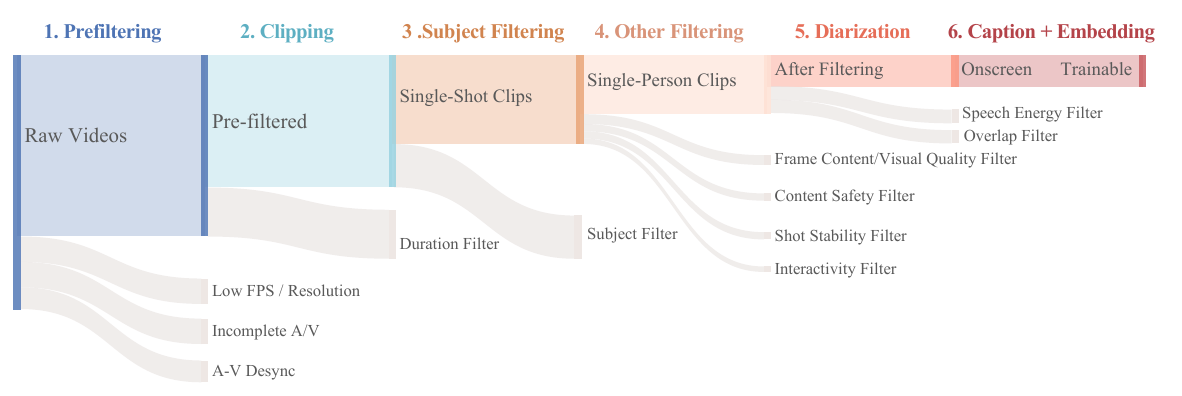}
    \caption{Data filtering pipeline for raw videos, showing progressive refinement through prefiltering, single-shot clipping, subject filtering, quality and safety checks, diarization, and caption generation. The final selected clips and their captions are embedded to form training data.}
    \label{fig:data-pipeline}
\end{figure}

As illustrated in Figure~\ref{fig:data-pipeline}, we build a progressive data processing pipeline that converts heterogeneous raw videos into high-quality, temporally coherent, and semantically annotated training clips. The pipeline first applies a prefiltering stage to discard technically unreliable videos, then segments the remaining data into single-shot clips, followed by subject-level filtering, quality and safety screening, speech diarization, and caption generation. This staged design jointly improves visual clarity, temporal stability, audio-visual consistency, and cross-modal interpretability, thereby yielding reliable training data for controllable character video generation. A detailed description of each stage is provided below.

\paragraph{Data Collection and Preprocessing.} 
The quality of the training data substantially affects both the training performance and the model's generalization capability. For this purpose, we collect and process a corpus of high-quality, diverse, and highly interactive single-person, single-shot video. The raw videos are drawn from two categories: livestream or talking-head videos, and high-quality footage from films and television dramas. The former is used mainly to learn fine-grained characteristics such as facial expressions, body movements, and lip synchronization, whereas the latter serves to improve the model's generalization and consistency across diverse shot angles, scenes, and visual styles. Before formally entering the pipeline, all data are first deduplicated and then pre-filtered according to metrics such as frame rate, resolution, audio-visual integrity, and audio-visual synchronization.

\paragraph{Data Clipping and Filtering.}
Raw videos are segmented into single-shot clips along shot boundaries, ensuring that the model learns continuous and coherent visual content. Long shots are further segmented, with cut points constrained so as not to fall in the middle of speech, ultimately yielding single-shot clips ranging from 3 to 60 seconds in duration. We evaluate the data comprehensively across multiple perspectives and levels, and apply strict filtering rules for cleaning, in order to guarantee the quality of the training data. In practice, we found that using expert models only exhibits clear limitations when assessing video data. For instance, face-detection models struggle to generalize to exaggerated or highly stylized 2D-animated subjects; some expert models are image-based and can only make judgments on sampled frames, rendering them prone to erroneous assessments owing to insufficient global information. We therefore introduce an omni model~\cite{xu2025qwen3,team2023gemini} to perform global semantic understanding over the complete video as a supplement. The omni model is used to generate semantic labels along multiple quality dimensions, including editing, subject, action, emotion, face, speech, scene, shot, and tone. The expert models and the Omni model together form a joint filtering system that possesses both global context awareness and local detail sensitivity. Specifically, we focus on the following aspects: \textit{(1) subject detection}, we ensure that each clip contains exactly one subject occupying a reasonable proportion of the frame; \textit{(2) frame cleanliness}, we filter out clips that contain information irrelevant to the visual content, such as watermarks, subtitles, and overlaid advertisements; \textit{(3) visual quality}, we use aesthetic and technical scoring to select clips that are clear, complete, and visually pleasing, while avoiding artifacts such as blur, jitter, and flicker; \textit{(4) content safety}, we filter out NSFW and other inappropriate content so as to prevent the model from learning harmful information; \textit{(5) shot stability}, we retain static shots or shots with slow motion in order to reduce the risk of shot drift during long-term generation; and \textit{(6) interactivity}, we require the subject in the frame to exhibit clear actions or behaviors, thereby enabling the model to learn meaningful motion information.

\paragraph{Speech Diarization.} High-fidelity lip synchronization can only be learned if the model observes visual performances that are consistent with the corresponding speech during training. To this end, we extract the speech component from the raw audio. We then apply voice activity detection and active speaker detection to annotate the timestamps of each speech segment together with its associated speaker. Based on the correspondence between the speaker and the on-screen subject, every segment is assigned to one of three categories: \textit{onscreen} (the speaker matches the person shown in the video), \textit{offscreen} (the speaker does not match any visible subject), and \textit{overlap} (multiple voices coincide within the segment). Clips containing overlapping segments are directly filtered out. During processing, we observed that the diarization model behaves unstably in scenarios involving singing or strong background music, frequently misassigning vocals to the music stem or introducing noticeable artifacts such as synthetic timbre and distortion. To mitigate this, we further introduce a heuristic rule that discards segments in which the speaker is vocalizing, but the speech energy proportion is too low. In this way, we preserve the cleanliness and reliability of the speech signal.

\paragraph{Data Captioning.} 
We annotate captions at two levels of granularity: full-clip captions and speech-aware chunk-level captions, which serve complementary roles. Clip-level captions provide a coherent global semantic anchor for the entire video, whereas speech-aware chunk-level captions align descriptions with their corresponding temporal intervals, offering fine-grained and temporally localized conditioning signals for controllable and interactive streaming generation. The annotation model produces structured natural-language descriptions covering a broad range of visual, auditory, and dialogue-related attributes, including subject appearance, actions, motion, emotion, scene context, camera language, cinematic properties, lighting, on-screen text, dialogue, sound effects, and background music. To improve annotation fidelity and reduce cross-modal hallucination, we adopt a dual-path strategy that decouples visual and auditory modalities. Visual attributes are inferred exclusively from video frames, while acoustic attributes are inferred exclusively from the audio track. Overall, the proposed structured annotation scheme improves the quality and consistency of multimodal representations while enabling more reliable annotation and more controllable generation.

\subsection{Training}

We aim to train a streaming video-audio generation model capable of producing temporally coherent video and synchronized audio under multimodal conditions. For clarity, we first introduce the notation used throughout the training framework.
\paragraph{Notation.}
Let ${\bm{v}}_0^i$ denote the clean video representation, and let ${\bm{a}}_0^i$ denote the clean audio representation associated with the $i$-th frame.
To formulate video-audio joint generation, we define the clean joint state as:
\begin{equation}
{\bm{x}}_0^i = [{\bm{v}}_0^i;{\bm{a}}_0^i],
\end{equation}
where $[\cdot;\cdot]$ denotes feature concatenation along the modality dimension.
For a video-audio sequence of length $N$, we denote the corresponding clean joint-state sequence as ${\bm{x}}_0^{1:N}=\{{\bm{x}}_0^1,\ldots,{\bm{x}}_0^N\}$.

The unified conditioning interface is denoted by ${\bm{c}}$, which includes speech, text prompts, and reference images.
For a given sequence, we represent the conditioning sequence as ${\bm{c}}^{1:\infty}=\{{\bm{c}}^1,{\bm{c}}^2,\ldots\}$, where ${\bm{c}}^i$ denotes the conditioning representation associated with the $i$-th video-audio state. 
The subscript $t_j \in \{t_0, \dots, t_T\}$ denotes the diffusion timestep of a noisy state, while $t_0=0$ corresponds to the clean state.
We use the hat notation $\hat{\cdot}$ to denote model-predicted quantities. 
In particular, $\hat{{\bm{x}}}_{0}^{i}$ denotes the predicted clean estimate of the $i$-th video-audio state. 
We use $f_\theta^{\mathrm{bi}}$ and $f_\theta$ to denote the bidirectional and causal denoising models, respectively.

Our training pipeline consists of three stages. 
In the first stage, we train a bidirectional video-audio generation model that serves as the foundation of our framework. 
In the second stage, we initialize the autoregressive model from the pretrained bidirectional model and further adapt it to a causal generation setting using a hybrid training strategy that unifies Teacher Forcing~\citep{gao2025ca2vdmefficientautoregressivevideo,jin2025pyramidal} and Diffusion Forcing~\cite{DF}.
This stage equips the model with stable multi-step autoregressive generation capabilities.
Finally, we employ  Distribution Matching Distillation (DMD)~\cite{yin2024one} with Phased Consistency Models (PCM)~\cite{wang2024phased} regularization to distill the autoregressive model into a few-step generator, significantly improving inference efficiency while maintaining generation quality.

\paragraph{Stage 1: Bidirectional Teacher Training.} 
We first train a bidirectional teacher on full video-audio sequences. 
For a finite sequence of $N$ frames, the bidirectional teacher is conditioned on the complete sequence ${\bm{c}}^{1:N}$ and is trained to denoise the corresponding joint latent states ${\bm{x}}_{0}^{1:N}$.
Its input and output at timestep $t_j$ can be written as:
\begin{equation}
\hat{{\bm{x}}}_{0}^{1:N}
=
f_\theta^{\mathrm{bi}}({\bm{x}}_{t_j}^{1:N},t_j,{\bm{c}}^{1:N}),
\label{eq:bidirectional_teacher}
\end{equation}
Here ${\bm{x}}_{t_j}^{1:N}$ denotes the noised version of joint video-audio states $1$ to $N$ at diffusion timestep $t_j$, ${\bm{c}}^{1:N}$ denotes the complete condition sequence, and $\hat{{\bm{x}}}_{0}^{1:N}$ denotes the clean-state estimate predicted from this noisy state. The training objective is defined as:
\begin{equation}
\mathcal{L}_{\mathrm{bi}}
=
\mathbb{E}_{({\bm{x}}_0^{1:N},{\bm{c}}^{1:N}),\,t_j}
\left[
\left\|
f_\theta^{\mathrm{bi}}({\bm{x}}_{t_j}^{1:N},t_j,{\bm{c}}^{1:N})
-
{\bm{x}}_{0}^{1:N}
\right\|_2^2
\right],
\label{eq:bi_loss}
\end{equation}

This stage establishes a high-quality generative prior for the subsequent causal adaptation and distillation.

\paragraph{Stage 2: Causal Teacher Training.}
We next initialize a causal teacher from the trained bidirectional teacher and adapt it to the streaming autoregressive setting. 
Specifically, we impose a causal attention mask on video-audio tokens, so that each target state can only access the causal condition context and the valid historical video-audio prefix. 

For each frame $i>1$, only the current joint state is denoised, conditioned on the available condition sequence ${\bm{c}}^{\leq i}$ and the historical video-audio prefix. 
The autoregressive denoising formulation can be written as:
\begin{equation}
\hat{{\bm{x}}}_{0}^{i}
=
f_\theta({\bm{x}}_{t_j}^{i},t_j,{\bm{c}}^{\leq i},{\bm{x}}_{\tau_j}^{<i},\tau_j),
\label{eq:ar_generation}
\end{equation}
where ${\bm{x}}_{t_j}^{i}$ is the noisy state of the current video-audio target at timestep $t_j$, ${\bm{c}}^{\leq i}$ denotes all conditions available up to frame $i$, ${\bm{x}}_{\tau_j}^{<i}$ denotes the historical video-audio prefix represented at noise level $\tau_j$, and $\hat{{\bm{x}}}_{0}^{i}$ is the clean-state estimate predicted from this noisy state.

To reduce the gap between training and inference, we adopt a hybrid training strategy that combines Teacher Forcing and Diffusion Forcing. 
For each training sample, we draw a sample from \(\mathrm{Bernoulli}(p)\) to choose between Teacher Forcing and Diffusion Forcing.
% For each training sample, we sample from $\mathrm{Bernoulli}(p)$ to decide whether it is trained with Teacher Forcing or Diffusion Forcing.

Under Teacher Forcing, the historical prefix is given by clean ground-truth video-audio states ${\bm{x}}_{0}^{<i}$, which is equivalent to setting $\tau_j=0$. This provides stable supervision for condition synchronization and motion consistency. Under Diffusion Forcing, the historical prefix is instantiated as noisy historical states ${\bm{x}}_{\tau_j}^{<i}$ with $\tau_j>0$, improving the model's robustness to imperfect prefixes during autoregressive rollout.
\begin{equation}
\mathcal{L}_{\mathrm{causal}}
=
\mathbb{E}_{({\bm{x}}_0^{1:N},{\bm{c}}^{1:N}),\,i,\,t_j,\,\tau_j}
\left[
\left\|
f_\theta({\bm{x}}_{t_j}^{i},t_j,{\bm{c}}^{\leq i},{\bm{x}}_{\tau_j}^{<i},\tau_j)
-
{\bm{x}}_{0}^{i}
\right\|_2^2
\right].
\label{eq:causal_loss}
\end{equation}
Through this adaptation, the causal teacher retains the generation quality of the bidirectional teacher while acquiring stable streaming generation capability.

\paragraph{Stage 3: DMD with PCM Regularization.}
Following the autoregressive warm-up stage, the model acquires a preliminary capability for multi-step autoregressive generation. 
However, generating videos still requires a large number of denoising steps for each autoregressive frame, resulting in substantial computational overhead and slow inference. 
To improve generation efficiency, we further apply DMD to compress the autoregressive generation process into only a few sampling steps. In our autoregressive setting, the causal generator predicts future states conditioned on the previously generated noisy prefix $\hat{{\bm{x}}}_{\tau_j}^{<i}$ and external conditions ${\bm{c}}^{1:\infty}$. 
By recursively applying the autoregressive generation process described in Equation.~(\ref{eq:ar_generation}), we obtain a generated video-audio sequence $\hat{\bm{x}}_0^{1:N}$. DMD optimizes the generator by following the gradient that minimizes the discrepancy between the generated and data distributions. Following~\citep{yin2024one}, the gradient of the DMD objective is given by

\begin{equation}
\nabla_{\theta}\mathcal{L}_{\mathrm{DMD}}
=
\mathbb{E}_{z,\,t}
\left[
w(t)
\left(
\nabla_{{\bm{x}}^{1:N}_{t}}
\log p_{\mathrm{data}}^{t}({\bm{x}}^{1:N}_{t})
-
\nabla_{{\bm{x}}^{1:N}_{t}}
\log p_{\theta}^{t}({\bm{x}}^{1:N}_{t})
\right)^{\!\top}
\frac{d {\bm{x}}^{1:N}_{t}}
{d\theta}
\right],
\end{equation}

where $\nabla_{{\bm{x}}_{t}}\log p_{\mathrm{data}}^{t}({\bm{x}}_{t})$
and
$\nabla_{{\bm{x}}_{t}}\log p_{\theta}^{t}({\bm{x}}_{t})$
denote the score functions of the data and generated distributions, respectively, and
$w(t)$ is the timestep-dependent weighting function.

Nevertheless, we observe that optimizing solely with the DMD objective frequently suffers from mode collapse, resulting in unstable generation behaviors such as camera drift, content degeneration, and temporal inconsistency.
To alleviate these issues, we introduce the distillation objective in PCM~\cite{wang2024phased} as an additional regularization mechanism during training. Specifically, we instantiate the distance function in the PCM objective using a perceptual feature distance, yielding

\begin{equation}
\mathcal{L}_{\mathrm{PCM}}
=
\mathbb{E}
\left[
\lambda(t_n)\,
d_{\mathrm{PERC}}
\!\left(
f_{\theta}({\bm{x}}^{1:N}_{t_{n+1}},t_{n+1}),
f_{\theta^-}(\tilde{{\bm{x}}}_{t_n}^{1:N},t_n)
\right)
\right],
\end{equation}

where $d_{\mathrm{PERC}}(\cdot,\cdot)$ denotes the perceptual feature distance, $f_{\theta^-}$ is the EMA student network, and $\lambda(t_n)$ is the timestep-dependent weighting function following PCM.

During training, we optimize the generator using a weighted combination of the DMD objective and the PCM consistency objective.

\subsection{Inference}

\subsubsection{Streaming Inference}
During inference, we employ sliding-window decoding to enable online autoregressive generation of arbitrarily long video-audio sequences under limited memory and computational resources. At the $i$-th autoregressive decoding step, attention is restricted to a fixed-length sliding window consisting of three components: (1) a persistent reference context, composed of the latent tokens extracted from the user-provided first frame together with the first generated video-audio state; (2) cached historical video-audio states retained within the sliding window; and (3) the current video-audio state being denoised.

The persistent reference context plays a role analogous to the sink token in large language models~\cite{xiao2024efficient} and the sink frame adopted by recent streaming video generation methods~\cite{yesiltepe2026infinity,yang2025longlive}, providing stable global conditioning throughout the generation process. Restricting attention to this fixed-length window satisfies the causal autoregressive constraint while keeping the per-step inference latency constant regardless of the generated sequence length. To further improve the efficiency and temporal stability of streaming inference, we employ Rotary Position Embedding~\citep{su2024roformer} (RoPE) Repositioning and TwinCache.

\paragraph{RoPE Repositioning}
To improve inference efficiency, we cache historical key-value features before applying RoPE~\cite{yesiltepe2026infinity,yi2025deep,li2026rolling,chen2026grounded,kim2026memrope}. As the sliding window advances, RoPE is applied to the cached features according to their updated relative positions in the current window, avoiding redundant recomputation of historical features. This strategy preserves cross-window positional consistency while ensuring that all visible positions remain within the positional range encountered during training.

\paragraph{TwinCache}
We introduce \textbf{TwinCache}, a stage-aware caching strategy that maintains complementary noisy and clean caches for historical video-audio states. The persistent reference context, consisting of the latent tokens extracted from the user-provided first frame together with the first generated video-audio state, is constructed only once and remains fixed throughout inference. In contrast, each subsequently generated video-audio state maintains two cached representations: a noisy cache extracted from a predefined denoising step and a clean cache obtained after the final denoising step.

During denoising, the model attends to the persistent reference context together with the cached historical representations. During intermediate denoising steps, the historical context is represented by the noisy cache, i.e., $\hat{\bm{x}}_{\tau_j}^{<i}$, whereas at the final denoising step it is replaced by the clean cache, i.e., $\hat{\bm{x}}_{0}^{<i}$.

Unlike clean representations, the residual noise retained in $\hat{\bm{x}}_{\tau_j}^{<i}$ preserves coarse temporal dynamics while suppressing the accumulation of high-frequency artifacts over long autoregressive sequences, acting as an implicit low-pass prior for temporal propagation~\cite{huang2025live,zheng2026causal}. Consequently, the noisy history cache provides stable temporal guidance without imposing strong appearance constraints. At the final denoising step, the clean history cache $\hat{\bm{x}}_{0}^{<i}$ is attended together with the persistent reference context to restore fine-grained visual details while maintaining temporal coherence and identity consistency~\cite{zeng2026lpm}. By decoupling temporal propagation from appearance refinement via stage-aware cache scheduling, TwinCache effectively balances long-term temporal consistency and visual fidelity in streaming video-audio generation.

\subsubsection{Inference Infrastructure}

Building on the inference configuration above, we deploy a hardware--software co-designed acceleration stack aligned with the technical route of TurboDiffusion~\cite{zhang2025turbodiffusion}, incorporating its key worker-side techniques, with TurboServe~\cite{jiang2026turboserve} as a reference for cluster-level streaming serving. By combining the above with our hardware-software co-design, we save per-step compute and memory in diffusion inference while preserving generation quality, enabling real-time inference on multiple GPUs. The main components are as follows.

\paragraph{Attention Acceleration.}
Attention is a major contributor to both latency and computational cost in the diffusion model~\cite{zhangefficient}. We accelerate attention with SageAttention~\cite{zhang2025sageattention,zhang2024sageattention2,zhang2025sageattention2++,zhang2025sageattention3,zhang2026sagebwd}, SpargeAttention~\cite{zhang2025spargeattention,zhang2026spargeattention2}, and Sparse-Linear Attention (SLA)~\cite{zhang2025sla,zhang2026sla2}. They significantly reduce attention latency while maintaining high generation quality.

\paragraph{Linear Layer Quantization.}
The choice of quantization granularity is critical to accuracy. We find that per-tensor and per-channel quantization are both vulnerable to outliers. A few extreme values stretch the quantization range, and the remaining values are therefore quantized with poor precision, which leads to noticeable quality loss. To address this, we implement custom per-block W8A8 quantized GEMM operators in CUDA. These operators reduce the inference-time memory footprint and significantly accelerate linear-layer computation, while maintaining high precision and generation quality.

\paragraph{Kernel Fusion.}
We fuse sequences of operators into custom Triton/CUDA kernels. Merging fine-grained kernels cuts host launch and synchronization overhead and avoids spilling intermediate tensors to global memory, reducing memory bandwidth pressure. Representative fusions include RMSNorm combined with selected elementwise operations.

\paragraph{CUDA Graph.}
Streaming video generation naturally exhibits two characteristics: individual operators have short computation times, and the inference process follows a relatively stable, repetitive execution pattern. Under these conditions, host-side kernel launch overhead accumulates and can become a bottleneck, leaving the GPU underutilized. We therefore adopt CUDA Graph: a fixed-structure subgraph is captured once into a CUDA graph and then replayed one or more times at each inference step. Each replay merges many independent kernel launches into a single graph launch, reducing redundant launch overhead and improving GPU utilization.

\paragraph{Multi-GPU Parallelism.}
To meet real-time latency targets within the available compute and memory budgets, we adopt Ulysses-style context parallelism~\cite{jacobs2023deepspeed} to distribute computation across multiple GPUs. Each GPU processes a slice of the sequence during the forward pass. Before and after attention, all-to-all collectives re-partition activations between sequence-parallel and head-parallel layouts, distributing both compute and activation memory across GPUs. We further optimize the underlying communication implementation to minimize its overhead.

\begin{table*}[t]
\centering
\small
\caption{
Quantitative comparison of different audio-driven avatar generation methods. 
Best results are highlighted in \textbf{bold}. 
Resolution and FPS/throughput are reported when publicly specified or measured in our test environment; otherwise marked as ``--''.
}
\label{tab:avatar_benchmark}
\setlength{\tabcolsep}{5.1pt}
\resizebox{\textwidth}{!}{
\begin{tabular}{lccccccc}
\toprule
\textbf{Model} &
\makecell{\textbf{Instruction}\\\textbf{Following}} &
\textbf{Real-Time} &
\textbf{Resolution} &
\makecell{\textbf{FPS /}\\\textbf{Throughput}} &
\textbf{CSIM} $\uparrow$ &
\textbf{Sync-D} $\downarrow$ &
\textbf{DOVER} $\uparrow$ \\
\midrule

OmniAvatar
& $\times$
& $\times$
& 480p
& --
& 0.8062
& 9.242
& 0.5476 \\

StableAvatar-1.3B
& $\times$
& $\times$
& 480p
& --
& 0.8358
& 11.18
& 0.5560 \\

Hallo3
& $\times$
& $\times$
& 480$\times$720
& --
& 0.7698
& 8.660
& 0.5313 \\

Wan2.2-S2V-14B
& $\times$
& $\times$
& 480p/720p
& --
& 0.7936
& 8.255
& 0.5510 \\

LiveAvatar
& $\times$
& \checkmark
& --
& --
& 0.8127
& 8.447
& 0.5639 \\

LemonSlice
& $\times$
& \checkmark
& 368$\times$560
& --
& 0.8407
& 7.921
& 0.5196 \\

HeyGen
& $\times$
& \checkmark
& --
& 25 FPS
& 0.9191
& 8.037
& 0.4864 \\

Kling Avatar 2.0
& \checkmark
& $\times$
& --
& --
& 0.8688
& 8.158
& 0.5406 \\

\midrule
\rowcolor[HTML]{F0F8FF}
\ours
& \checkmark
& \checkmark
& 540p (960$\times$540)
& \textbf{42 FPS}
& \textbf{0.9192}
& \textbf{7.8470}
& \textbf{0.5660} \\

\bottomrule
\end{tabular}
}
\end{table*}

\section{Evaluations}

\subsection{Setup}

Standard public benchmarks such as HDTF~\cite{zhang2021flow} evaluate audio-driven avatar generation under controlled settings, but they do not fully capture whether a model can follow action instructions and produce natural motion in real-time interaction. We therefore introduce Vidu-StreamBench, an in-house benchmark for real-time interactive avatar generation, while also reporting HDTF results as a standard public reference. We further measure inference efficiency on RTX 5090 GPUs to verify the real-time capability of \ours.

\paragraph{Benchmarks.}
Vidu-StreamBench contains 500 samples, each composed of an action instruction, a reference first frame, and an audio clip. The benchmark covers diverse action commands, reference image styles, speaker attributes, emotions, and application scenarios. It is designed to test whether a model can generate avatars that move naturally, remain stable over time, preserve identity, follow user instructions, and maintain coherent audio-video behavior in realistic streaming settings. We conduct pairwise A/B preference tests against three leading commercial systems: HeyGen~\cite{heygen2026avatar}, LemonSlice~\cite{lemonslice2026studio}, and Kling Avatar 2.0~\cite{klingteam2025klingavatar}.

For public benchmark evaluation, we use HDTF and compare \ours with the same commercial systems, as well as recent open-source models, including Wan2.2-S2V-14B~\cite{gao2025wan}, LiveAvatar~\cite{huang2025live}, OmniAvatar-1.3B~\cite{gan2025omniavatar}, Hallo3~\cite{cui2025hallo3}, and StableAvatar-1.3B~\cite{tu2025stableavatar}. In detail, we report CSIM~\cite{deng2019arcface}, Sync-D~\cite{prajwal2020lip}, and DOVER~\cite{wu2023exploring} to measure identity preservation, audio-visual synchronization, and perceptual video quality.

\begin{figure}[h]
\centering
\includegraphics[width=1.0\textwidth]{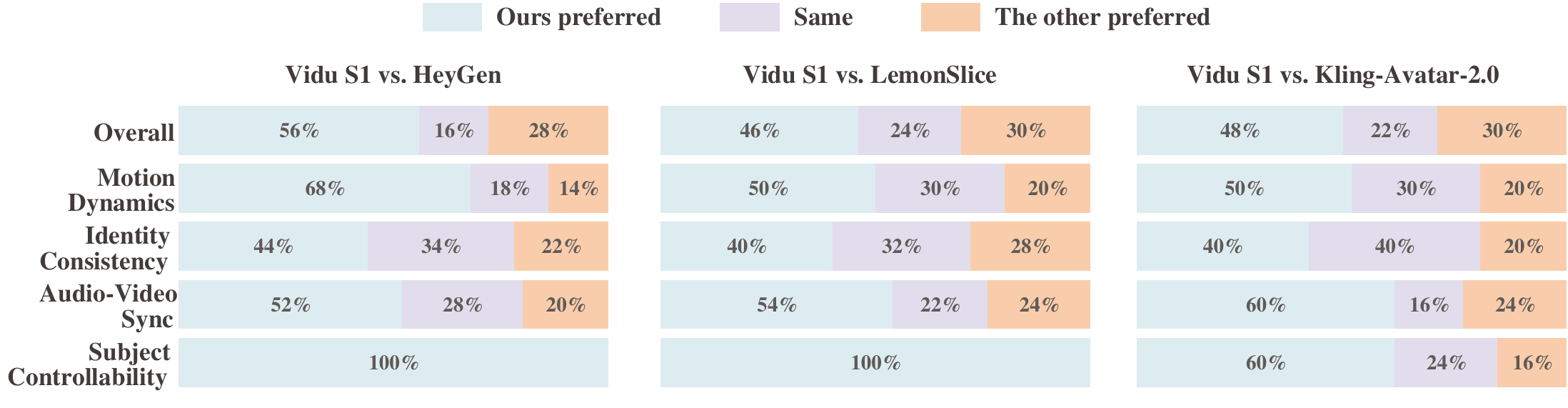}
\caption{
Human preference evaluation of \ours versus HeyGen, LemonSlice, and Kling-Avatar-2.0 on Vidu-StreamBench.
}
\label{fig:streambench_human_eval}
\end{figure}

\subsection{Results}

\subsubsection{Benchmark Evaluation}

\paragraph{Analysis.}
Figure~\ref{fig:streambench_human_eval} summarizes the pairwise preference results on Vidu-StreamBench. 
Across comparisons with leading commercial avatar systems, \ours is consistently preferred overall.
These results suggest that \ours is more effective at translating interactive user instructions into visible avatar behavior, including natural body motion, gestures, head movement, and controllable action execution, rather than only producing passive speech-driven facial animation.

The advantage is especially evident in subject controllability, where \ours achieves a 100\% preference rate against HeyGen and LemonSlice. This indicates that human raters favored \ours when the task required the avatar to follow explicit action instructions. \ours also shows consistent gains in motion dynamics and audio-video synchronization, further supporting its suitability for real-time interactive avatar applications.

\begin{figure*}[t]
\centering
\includegraphics[width=\textwidth]{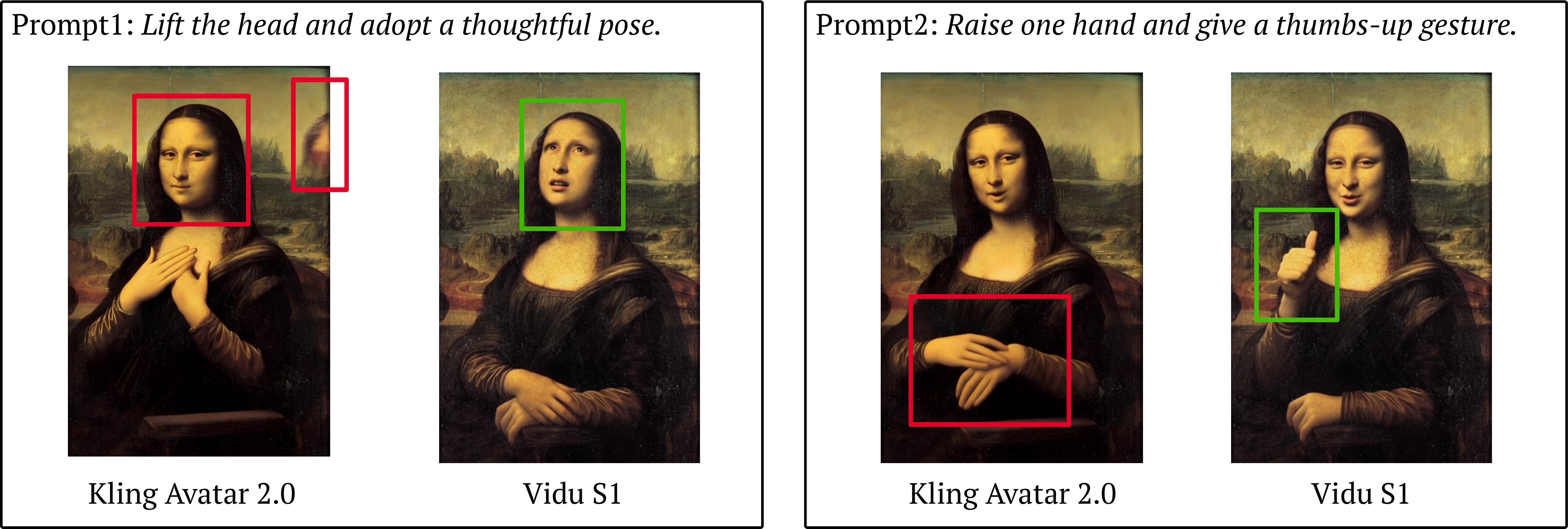}
\caption{
Qualitative comparison of instruction following and visual consistency.
\textcolor[rgb]{0.55,0.0,0.0}{
Red boxes} highlight failure cases from the compared method: in Prompt 1, the subject does not raise the head and noticeable background artifacts appear; in Prompt 2, the thumbs-up gesture is not correctly performed.
\textcolor[rgb]{0.0,0.45,0.2}{
Green boxes} highlight the results of \ours: in Prompt 1, the model correctly follows the instruction by raising the head and adopting a thoughtful pose; in Prompt 2, it successfully performs the thumbs-up gesture.
}
\label{fig:visual_compare}
\end{figure*}

\paragraph{Visualization.}
Figure~\ref{fig:visual_compare} compares \ours with Kling Avatar 2.0 on Vidu-StreamBench. Given the same reference image, audio, and action instruction, \ours demonstrates stronger visual stability and instruction following, leading to more engaging interactive behavior and better suitability for long-horizon use.

Table~\ref{tab:avatar_benchmark} reports complementary quantitative results on HDTF. \ours achieves leading performance across identity preservation, audio-visual synchronization, and perceptual quality, while also supporting real-time generation and instruction following.

\subsubsection{Real-Time Interactive Generation Capability}

Beyond benchmark quality, real-time interactive video generation requires low-latency inference, continuous user control, and stable streaming behavior. \ours is designed for this setting: it supports online prompt-level interaction and uses speech as an explicit control signal, allowing users to control the generated video during the generation process rather than specifying all conditions beforehand.

As shown in Table~\ref{tab:avatar_benchmark}, \ours generates 540p video with a 3-step configuration and reaches an average throughput of 42 FPS on RTX 5090 GPUs. This exceeds the 30-FPS real-time playback threshold, demonstrating the practical feasibility of \ours for latency-sensitive applications such as live conversation, virtual hosts, entertainment avatars, and educational agents. Importantly, this inference efficiency is achieved while preserving the strong identity consistency, audio-visual synchronization, and perceptual quality demonstrated on HDTF, showing that \ours delivers real-time performance without sacrificing avatar generation quality.
\section{Conclusion}
We present \ours, a real-time interactive video generation model for voice-controlled digital characters. \ours enables users to guide video content at any moment through voice instructions and supports infinite-length generation with stable visual quality and subject identity. With an efficient inference and serving pipeline, it produces 540p videos at up to 42 FPS on consumer GPUs. Experiments show that \ours achieves the best overall performance while meeting real-time inference requirements. These results highlight the potential of \ours for practical, controllable, and personalized video generation. We hope this work provides a useful step toward real-time generative video systems that are realistic, consistent, and easy to control.

% \clearpage

\bibliographystyle{unsrt}
\bibliography{main}

\end{document}